\documentclass[runningheads]{llncs}
\usepackage{graphicx}
\usepackage{epsfig}
 \usepackage{xspace}
\usepackage{tikz}
\usepackage{subfigure}
\usepackage{comment}
\usepackage{mathrsfs}
\usepackage{amsmath,amssymb} 
\usepackage{color}
\usepackage{multirow}
\usepackage[linesnumbered,boxed,ruled,commentsnumbered]{algorithm2e}

\begin{document}
\pagestyle{headings}
\mainmatter
\def\ECCVSubNumber{3671}  

\title{E-Graph: Minimal Solution for Rigid Rotation with Extensibility Graphs} 

\titlerunning{ECCV-22 submission ID \ECCVSubNumber} 
\authorrunning{ECCV-22 submission ID \ECCVSubNumber} 
\author{Anonymous ECCV submission}
\institute{Paper ID \ECCVSubNumber}

\titlerunning{Extensibility Graph}
%
\author{Yanyan Li\inst{1,2}\orcidID{0000-0001-7292-9175} \and
Federico Tombari\inst{1,3}\orcidID{0000-0001-5598-5212}}
\authorrunning{Y. Li and F. Tombari}
%
\institute{Technical University of Munich, Munich, Germany\and
Meta-Bounds Tech, Shenzhen, China \\ 
\and
Google, Zurich, Switzerland \\
\email{yanyan.li@tum.de, tombari@in.tum.de}}
\maketitle

\begin{abstract}
Minimal solutions for relative rotation and translation estimation tasks have been explored in different scenarios, typically relying on the so-called co-visibility graphs. However, how to build direct rotation relationships between two frames without overlap is still an open topic, which, if solved, could greatly improve the accuracy of visual odometry.
In this paper, a new minimal solution is proposed to solve relative rotation estimation between two images without overlapping areas by exploiting a new graph structure, which we call Extensibility Graph (E-Graph). 
Differently from a co-visibility graph, high-level landmarks, including vanishing directions and plane normals, are stored in our E-Graph, which are geometrically extensible. 
Based on E-Graph, the rotation estimation problem becomes simpler and more elegant, as it can deal with pure rotational motion and requires fewer assumptions, e.g.\ Manhattan/Atlanta World, planar/vertical motion.
Finally, we embed our rotation estimation strategy into a complete camera tracking and mapping system which obtains 6-DoF camera poses and a dense 3D mesh model. 
Extensive experiments on public benchmarks demonstrate that the proposed method achieves state-of-the-art tracking performance. 
\end{abstract}

\section{Introduction}
Camera pose estimation is a long-standing problem in computer vision as a key step in algorithms for visual odometry, Simultaneous Localization and Mapping (SLAM) and related applications in robotics, augmented reality, autonomous driving (to name a few). 
\begin{figure}
    \centering
    \subfigure[Dense scene reconstruction]
    {\includegraphics[width=0.40\textwidth]{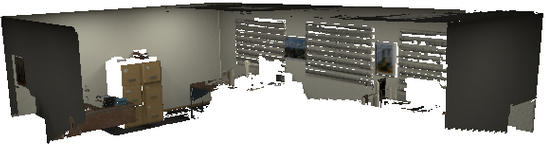}}
    \subfigure[Sparse scene reconstruction]
    {
    \includegraphics[width=0.40\textwidth]{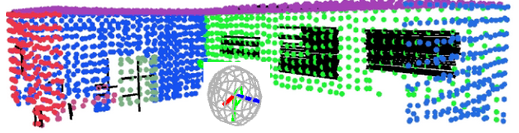}}
    \subfigure[Covisibility graph]{
     \includegraphics[scale=0.50]{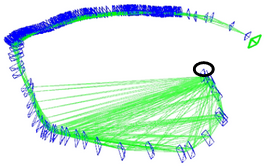}}
    \subfigure[Extensibility graph]{
      \includegraphics[scale=0.50]{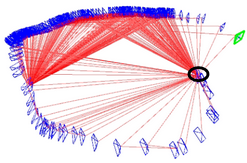}}
    \caption{Dense (a) and sparse (b) scene reconstruction of the office-room scene from the ICL dataset~\cite{handa:etal:ICRA2014} obtained by the proposed method. (c) and (d): keyframes (in blue) and connected frames are linked with green and red lines, respectively, to build up the proposed covisibility and extensibility graphs. The black ellipses denote the start points of the camera trajectory.}
    \label{fig:teaser}
\end{figure}
As part of the camera pose estimation problem, the minimal case~\cite{zhao2019minimal} provides an estimate of whether the problem can be solved and how many elements are required to obtain a reliable estimate. According to the input data type and scenarios, different solutions~\cite{andrew2001multiple,nister2004efficient,kim2018indoor,guan2020minimal} were proposed, most of which became very popular in the computer vision and robotic community, such as the seven-point~\cite{andrew2001multiple} and five-point~\cite{nister2004efficient} approaches. A typical limitation of traditional pose estimation solutions based on the minimal case~\cite{andrew2001multiple,nister2004efficient,salaun2016robust,guan2020minimal} is that both rotation and translation estimation rely on the co-visibility features between two frames, this having as a consequence that the length of an edge between two nodes is often relatively short. Therefore, tracking errors tend to accumulate easily based on a frame-to-frame or frame-to-keyframe strategy. To solve this issue, more advanced tracking systems~\cite{murORB2,campos2020orb} with optimization solutions, including local and global bundle adjustment approaches, were exploited to refine poses from minimal solutions. Loop Closure is a common algorithm used in feature-based~\cite{murTRO2015} and direct~\cite{gao2018ldso} methods to remove drift. However, it also requires the camera to revisit the same place, which is a limiting assumption in many scenarios. 

Compared with point features, lines and planes require more computation to be extracted and described. Early multi-feature SLAM systems~\cite{gomez2019pl} use them to increase the number of features to combat low-textured scenes. After that, co-planar, parallel and perpendicular relationships were explored~\cite{zhang2019point,li2020co,Li2021PlanarSLAM} to add more constraints in the optimization module, still following a similar tracking strategy as ORBSLAM~\cite{mur2015orb} or DSO~\cite{wang2017stereo} for the initial pose estimation. 

Different to the tightly coupled estimation strategy, some works~\cite{zhou2016divide} proposed to decouple the 6-DoF pose estimation into rotation and translation estimation aiming to achieve a more accurate rotation estimation, based on the idea that pose drift is mainly caused by the rotation component~\cite{kim2018linear}.  
At the same time, based on an estimated rotation matrix~\cite{salaun2016robust}, only two points are required to compute the translation motion, leading to more robustness in low-textured regions.

The Manhattan World (MW)~\cite{zhou2016divide} and Atlanta World (AW)~\cite{joo2020linear} assumptions introduce stronger constraints since they require a single orthogonal scene, or a scene with a unified vertical direction. Unlike loop closure that removes drift by detecting trajectory loops, the assumption of MW and AW is introduced for indoor tracking scenarios~\cite{li2020structure,kim2018linear} to improve the accuracy of camera pose estimation, since most indoor artificial environments follow this assumption. MW and AW improve accuracy when the main structure of the scene has orthogonal elements 
However, since this assumption requires the observation of vertical/orthogonal environmental features (such as straight lines or planes), the SLAM system using this method is also limited in the types of scenarios it can be successfully applied to.

In this paper we propose a rigid rotation estimation approach based on a novel graph structure, which we dub Extensibility Graph (E-Graph), for landmark association in RGB-D data. Our approach is designed to reduce drift and improve the overall trajectory accuracy in spite of loop closure or MW/AW assumptions. Benefiting of E-Graph, the drift-free rotation estimation problem is simplified to the alignment problem of rotating coordinate systems. Importantly, our rotation step does not need overlaps between two frames by making use of vanishing directions of lines and plane normals in the scene, hence can relate a higher number of keyframes with respect to standard co-visibility graphs, with benefits in terms of accuracy and robustness in presence of pure rotational motions.

In addition, we develop a complete tracking and dense mapping system base on the proposed E-Graph and rotation estimation strategies, which we demonstrate to outperform state-of-the-art SLAM approaches~\cite{Li2021PlanarSLAM,yunus2021manhattanslam,murORB2,campos2020orb}. 
To summarize, the main contributions of this paper are as follows:
i) a new perspective for reducing drift is proposed based on our novel graph structure, E-Graph, which connects keyframes across long distances;
ii) a novel drift-free rotation alignment solution between two frames without overlapping areas based on E-Graph; 
iii) a complete SLAM system based on the two previous contributions to improve robustness and accuracy in pose estimation and mapping. 
The proposed approach is evaluated on common benchmarks such as ICL~\cite{handa:etal:ICRA2014} and TUM-RGBD~\cite{sturm2012benchmark}, demonstrating an improved performance compared to the state of the art.

\section{Related work}\label{related_work}
By making the assumption of planar motion~\cite{guan2020minimal}, two-view relative pose estimation is implemented based on a single affine correspondence.
Point features are common geometric features used in VO and SLAM~\cite{campos2020orb} systems. To remove the drift from point-based front ends, different types of back ends are explored in tracking methods. Loop closing is an important module to remove drift, which happens when the system recognizes that a place~\cite{6202705,mei2010closing} has been visited before. After closing the loop, associated keyframes in the covisibility graph will be adjusted. Benefiting of loop closure and optimization modules, ORB-SLAM series~\cite{murORB2,campos2020orb} organize the keyframes efficiently, which provides robust support for tracking tasks. Different from sparse point features used in ORB-SLAM, BAD-SLAM~\cite{8954208} implements a direct bundle adjustment formulation supported by GPU processing.

However, in indoor environments, to cover texture-less regions that have few point features, more geometric features are merged into the front end of systems. At the early stage, methods build re-projection error functions for lines and planes. CPA-SLAM~\cite{7487260} makes use of photometric and plane re-projection terms to estimate the camera pose. Based on estimated camera poses, detected planes are merged together with a global plane model. Similar to our method, CPA-SLAM and KDP-SLAM~\cite{7989597} can build constraints between non-overlapping frames. However those constraints are used to build heavy optimization targets instead of improving the efficiency. Furthermore, the relationship between parallel lines (vanishing points) and perpendicular planes is explored in~\cite{8793716,7001715}. Based on the regularities of those structural features, they obtain a more accurate performance. Instead of exploring the parallel/perpendicular relationships between lines/planes, ~\cite{rosinol2019incremental,li2020co} make use of constraints between co-planar points and lines in the optimization module. 

Those regularities aim to build constraints between local features, \cite{kim2018linear,Li2021PlanarSLAM} introduce global constraints by modeling the environment as a special shape, like MW and AW. The MW assumption is suitable for a cuboid scenario, which is supposed to be built by orthogonal elements. Based on this assumption, those methods estimate each frame's rotation between the frame and the Manhattan world directly, which is useful to avoid drift between frames in those scenes. 
L-SLAM~\cite{kim2018linear} groups normal vectors of each pixel into an orthogonal coordinate by projecting them into a Gaussian Sphere~\cite{zhou2016divide} and tracks the coordinate axes to compute the relative rotation motion. Similar to the main idea of L-SLAM, ~\cite{kim2018indoor} provides a RGB-D compass by using a single line and plane. Since the line lies on the plane, the underlying assumption of the system is the MW-based rotation estimation method. However, the limitation of this strategy is also very obvious, that it works only in Manhattan environments. Based on ORB-SLAM2~\cite{murORB2}, Structure-SLAM~\cite{li2020structure,Li2021PlanarSLAM} merges the MW assumption with keyframe-based tracking, to improve the robustness of the system in non-MW indoor scenes, which refine decoupled camera pose by using a frame-to-model strategy. Compared with MW-based tracking methods, our approach is less sensitive to the structure of environments.

\section{Minimal case in orientation estimation}
Commonly, the 6-DoF Euclidean Transform $\mathit{T}\in SE(3)$ defines motions as a set of rotation $\mathit{R}\in SO(3)$ and translation $\textbf{t} \in \mathbb{R}^3$. Based on point correspondences, camera pose estimation can be defined as, 
\begin{equation}\label{eq:basic}
    \textbf{P}^{'} = R\textbf{P}+\textbf{t}
\end{equation}
where $ \textbf{P}^{'}$ and $ \textbf{P}$ are 3D correspondences, and $[R, \textbf{t}]$ defines the relative motion between two cameras. For monocular sensors, their image normalized representations are $ \textbf{X}_c^{'}$ and $ \textbf{X}_c$,   
\begin{equation}\label{eq:rt}
     \textbf{X}_c^{'} = \alpha(R  \textbf{X}_c + \gamma  \textbf{t})
\end{equation}
where $\alpha$ and $\gamma$ are depth-related parameters. After multiplying (\ref{eq:rt}) by $ \textbf{X}_c^{'T}[ \textbf{t}]_x$, we can obtain the classic essential matrix equation,
\begin{equation}\label{eq:basic}
     \textbf{X}_c^{'T}E \textbf{X}_c = 0
\end{equation}
where $E=[ \textbf{t}]_xR$ and $[ \textbf{t}]_x$ is the skew symmetric matrix formed by $ \textbf{t}$. 

For RGB-D sensors, the task is simplified since the absolute depth information is directly provided by sensors. Equation~(\ref{eq:basic}) can be solved by using 3 non-collinear correspondences only~\cite{pomerleau2015review}, although the distance between two frames is supposed to be kept small to extract enough correspondences. 

\subsection{Minimal solution for rotation}\label{rotation}

\begin{figure}
    \centering
    \includegraphics[width=\linewidth]{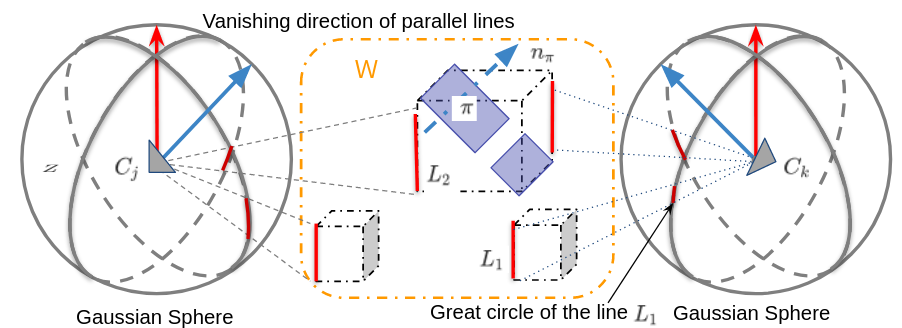}
    \caption{Minimal case of rotation estimation in EG}
    \label{fig:relative_rotation}
\end{figure}

Different from traditional methods based on co-visibility graphs, the proposed method decouples rotation and translation estimation into two separate stages. Moreover, the rotation estimation task does not require feature correspondences.
As shown in Figure~\ref{fig:relative_rotation}, non-parallel direction vectors $\textbf{v}_m, m\in [0,1, \dots, n]$ are detected in the camera coordinate $C_j$, where $\textbf{v}_m^j = [_xv_m^j, _yv_m^j, _zv_m^j]^T$. In Euclidean 3D space, the size of a finite and linearly independent set of vectors is less then four. According to the Gram-Schmidt orthogonalization process, we can obtain an orthogonal set $S = [\textbf{u}_0, \textbf{u}_1, \textbf{u}_2 ]$,
\begin{equation}
\begin{array}{cl}
     \textbf{u}_0 & = \textbf{v}_0^j  \\
     \textbf{u}_1 & = \textbf{v}_1^j - proj_{[\textbf{v}_0^j]}(\textbf{v}_1^j) \\
     \textbf{u}_2 & = \textbf{v}_2^j - proj_{[\textbf{v}_0^j]}(\textbf{v}_2^j)-proj_{[\textbf{v}_1^j]}(\textbf{v}_2^j)
\end{array}
\end{equation}
by using the projection operator $proj_{[\textbf{u}]}(\textbf{v})=\frac{<\textbf{u},\textbf{v}>}{<\textbf{u},\textbf{v}>}\textbf{u}$, where $<\textbf{u},\textbf{v}>$ shows the inner product of the vectors \textbf{u} and \textbf{v}.  
Furthermore, we obtain the normalized vectors $\textbf{e}_0$, $\textbf{e}_1$ and $\textbf{e}_2$ via $\textbf{e}_m=\frac{\textbf{u}_m}{||\textbf{u}_m||}$. 

For the Euclidean space $\mathbb{R}^3$, the relevant orthonormal basis set based on the detected direction vectors is $(\textbf{e}_0, \textbf{e}_1, \textbf{e}_2)$. In the $j^{th}$ camera coordinate, the orthonormal set is detected as $(\textbf{e}_0, \textbf{e}_1,\textbf{e}_2)$, while $(\textbf{e}_0^*, \textbf{e}_1^*,\textbf{e}_2^*)$ in the $k^{th}$ camera coordinate. 

Therefore, from the perspective of the orthonormal set, those $j^{th}$ and $k^{th}$ coordinates are represented as $[\textbf{e}_0, \textbf{e}_1,\textbf{e}_2]^{T}$ and
$[\textbf{e}_0^*, \textbf{e}_1^*,\textbf{e}_2^*]^{T}$, respectively. 

Given $\left[\begin{array}{c}
     \textbf{e}_0^T  \\
     \textbf{e}_1^T  \\
     \textbf{e}_2^T
\end{array}\right][\textbf{e}_0, \textbf{e}_1,\textbf{e}_2]$ is the identity matrix, the matrix $[\textbf{e}_0, \textbf{e}_1,\textbf{e}_2]$ is an orthogonal matrix and the columns of $[\textbf{e}_0, \textbf{e}_1,\textbf{e}_2]^T$ are orthonormal vectors as well, which can be used to build the orthonomal basis set of the $j^{th}$ camera coordinate. Therefore, in $\mathbb{R}^3$ an arbitrary vector $\textbf{x}$ can be represented by two orthonormal sets, $ (\textbf{e}_0, \textbf{e}_1, \textbf{e}_2)^{T}$ and $(\textbf{e}_0^*, \textbf{e}_1^*, \textbf{e}_2^*)^{T}$, independently, 
\begin{equation}
    \begin{array}{cl}
         \textbf{x} &  = (\textbf{e}_0, \textbf{e}_1, \textbf{e}_2)^{T} ( x_0, x_1, x_2)^T \\
         & = (\textbf{e}_0^*, \textbf{e}_1^*, \textbf{e}_2^*)^{T} ( x_0^*, x_1^*, x_2^*)^T
    \end{array}
\end{equation}

Finally, $(x_0, x_1, x_2)^T =(\textbf{e}_0, \textbf{e}_1, \textbf{e}_2)  (\textbf{e}_0^{*}, \textbf{e}_1^{*}, \textbf{e}_2^{*})^T( x_0^*, x_1^*, x_2^*)^T$ where the rotation motion $R_{c_jc_k}$ from camera $k$ to camera $j$ is $[\textbf{e}_0, \textbf{e}_1, \textbf{e}_2]  [\textbf{e}_0^{*}, \textbf{e}_1^{*}, \textbf{e}_2^{*}]^T$.

\paragraph{\textbf{Two-Observation case}.} In the spatial case where two linearly independent direction vectors are detected, $\textbf{u}_2$ can be achieved by the cross product process of $\textbf{u}_0$ and $\textbf{u}_1$. Obviously, the new set $[\textbf{u}_0, \textbf{u}_1, \textbf{u}_0\times \textbf{u}_1]$ maintains the orthogonal property, which is the minimal solution for relative pose estimation problems.

\paragraph{\textbf{Orthogonal-Observation case}.}
As discussed in Section~\ref{related_work}, the MW assumption is enforced mostly by SLAM/VO methods designed to work indoor~\cite{kim2018indoor,kim2018linear,li2020structure,yunus2021manhattanslam}, achieving particularly good results when the MW assumption holds. When the observation vectors $\textbf{v}_m^j$ are orthogonal, the projection operation between different vectors is zero and the proposed method degenerates to a multi-MW case, 
\begin{equation}
\begin{array}{cl}
    R_{c_jc_k} & = R_{c_jM_i}R_{c_kM_i}^T  \\
     & = [\frac{\textbf{v}_0^j}{||\textbf{v}_0^j||}, \frac{\textbf{v}_1^j}{||\textbf{v}_1^j||}, \frac{\textbf{v}_2^j}{||\textbf{v}_2^j||}][\frac{\textbf{v}_0^k}{||\textbf{v}_0^k||}, \frac{\textbf{v}_1^k}{||\textbf{v}_1^k||}, \frac{\textbf{v}_2^k}{||\textbf{v}_2^k||}]^T.
\end{array}
\end{equation}

For single-MW scenarios, a global orthogonal set can be obtained by every frame, therefore $R_{c_jw}$, from world to camera $C_j$, can be computed by $R_{c_jM}R_{c_0M}^T$, here $R_{c_0w}$ is an identity matrix.

Compared with the visual compass~\cite{kim2018indoor} method making use of a combination of line and plane features from MW~\cite{kim2018linear} to estimate camera rotation, our graph is more robust and flexible. Furthermore, compared to~\cite{salaun2016robust} that generates four rotation candidates after aligning two frames' vanishing points, our method not only leverages plane features, but also solves the ambiguity regarding the directions of the vanishing points~\cite{salaun2016robust}.

After the relative rotation pose estimation step between two frames, in case of no overlap between them, we need to make use of their neighboring frames to compute translation vectors. Note that only two correspondences are required in translation estimation by making use of Equation~\ref{eq:basic}, which is particularly suited to deal with scenes and environments characterized by different texture types compared to traditional approaches~\cite{murORB2,campos2020orb}.



\section{Extensibility Graph (E-Graph)} 
\begin{figure}
    \centering
\subfigure[]{
    \begin{minipage}[b]{0.2\linewidth}
    \includegraphics[height=2cm]{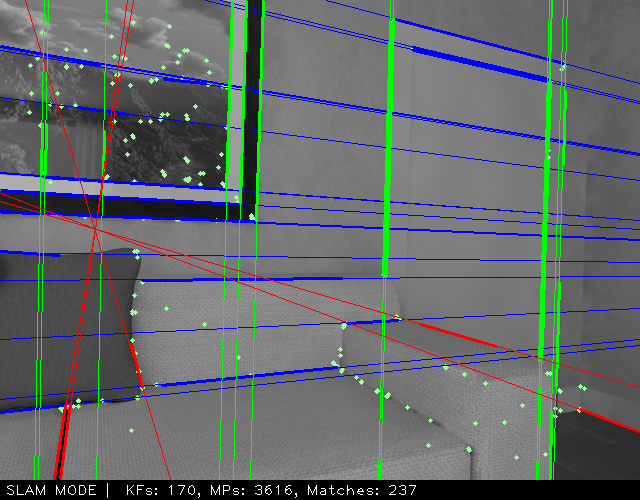}
    \includegraphics[height=2cm]{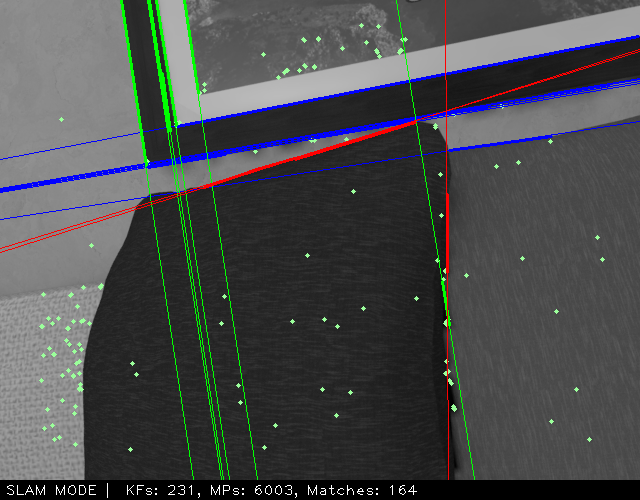}
    \end{minipage}}
\subfigure[]{
    \begin{minipage}[b]{0.2\linewidth}
    \includegraphics[height=2cm]{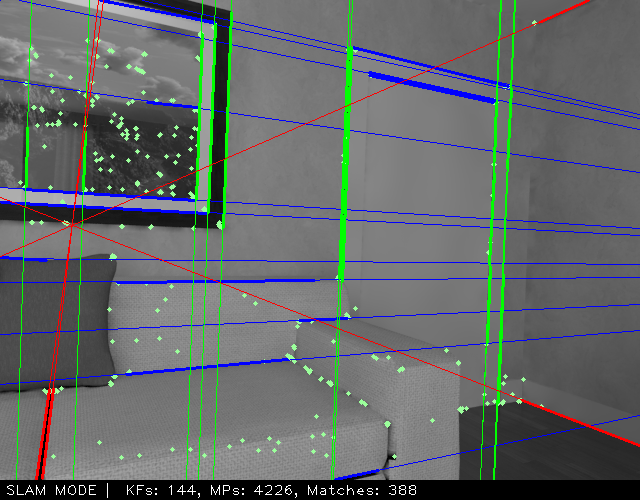}
    \includegraphics[height=2cm]{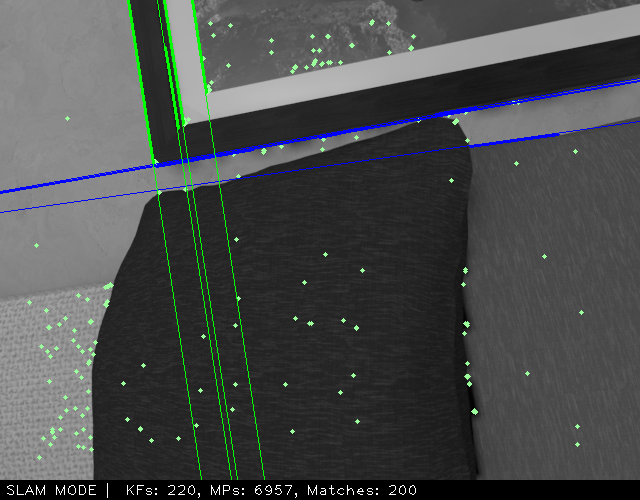}
    \end{minipage}}
    \subfigure[]{
     \begin{minipage}[b]{0.5\linewidth}
    \includegraphics[height=4cm]{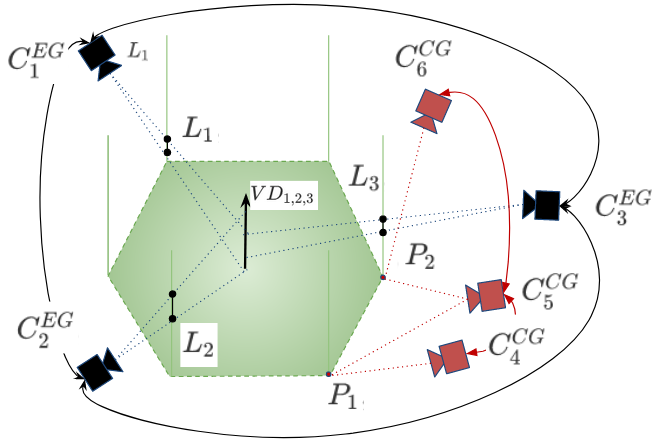}
    \end{minipage}}
    \caption{Vanishing point detection and Rotation connection examples. (a) Detection results of J-Linkage. (b) Refined results by our system. (c) E-Graph (black) and co-visibility graph (red).}
    \label{fig:graphs}
\end{figure}
As shown in Figure~\ref{fig:graphs}(c), the E-Graph method builds rotation connections (edges) between frames $[C_1^{EG}, C_2^{EG}, C_3^{EG}]$ that share global directions instead of any low-level correspondences (like points and lines). At the same time, no connection between $C_4^{CG}$ and $C_6^{CG}$ can be made since these frames have no co-visible features within the co-visibility graph. The proposed connection strategy will be detailed in the following subsections. 

\subsection{Landmarks from a RGB-D frame}
Similar to the co-visibility graph, the proposed graph is also a topological representation of scenes. The difference is that the proposed graph is built based on the scene structure rather than on overlapping parts between frames. The distance between connected frames in a co-visibility graph tends to be small (see Figure~\ref{fig:graphs}) since two frames that are distant from each other rarely overlap, leading to the pose of the current frame being estimated based on the last frame or last keyframes only. The issue can be alleviated by using global bundle adjustment and loop closure modules, although they bring in intensive computation and trajectory constraints (e.g.\ need to re-visit a certain area). 

In our graph $\mathscr{G}=[\mathscr{N}_c,\mathscr{N}_{lm},\mathscr{E}]$ frames and landmarks are regarded as nodes $\mathscr{N}_c$ and $\mathscr{N}_{lm}$ respectively, while $\mathscr{E}$ represents the edges among connected frames. Note that landmarks are border-less planes and vanishing directions, e.g.\ $\mathit{VD}_{1,2,3}$, of lines detected in multiple views. In particular, an edge is established between two frames every time two or more structural elements are matched across them. 

\paragraph{\textbf{Features and landmarks}.}\label{sec:landmarks}
Vanishing directions are estimated from parallel lines detected by a joint 2D-3D process, where LSD~\cite{von2010lsd} is used to extract 2D line features from RGB images. Meanwhile, AHP~\cite{feng2014fast} is carried out to extract plane features from depth maps.

Firstly, as shown in Figure~\ref{fig:graphs}(a), we make use of the J-Linkage algorithm to classify detected 2D lines into different groups of parallel lines as described in~\cite{tardif2009non}. However, there are still outliers left based on the 2D process. To solve this issue, we take advantage of depth maps to check the directions of lines in each group by using RANSAC to detect the best direction vector $\mathit{VD}_n$ to represent the group $S_n$. 



As for planar landmarks, we make use of the Hessian ($\pmb{\pi}=(\textbf{n}^{\pi},d^{\pi})$) to represent a plane detected from the $i^{th}$ frame, where $\textbf{n}^{\pi}$ denotes the normal vector and $d^{\pi}$ represents the distance between the camera center and this plane, which is transferred to world coordinates via the initial pose $T_{wc_i}$. 

\subsection{Data Association}\label{association}
After generating vanishing directions and planes, we now explain how to initialize and update them. 
\paragraph{\textbf{Initialization}.}
Combined with the first keyframe $\mathit{Kf}_0$, detected planes and optimized vanishing directions are used to initialize the E-Graph. The camera pose $T_0$ of $\mathit{Kf}_0$ is set as the world coordinate for landmarks in the E-Graph. Planes $\pmb{\pi}_i$ measured by $\mathit{Kf}_0$ are transferred to the graph directly as,
\begin{equation}
    \mathscr{G}_0 = [\mathscr{N}_{c_0},\mathscr{N}_{lm_0},\mathscr{E}_0] 
\end{equation}
where $\mathscr{N}_{c_0}$ is $\mathit{Kf}_0$ and $\mathscr{E}_0$ has no edges yet. $\mathscr{N}_{lm_0}$ contains $[\pmb{\pi}_i, \mathit{VD}_i, \mathit{PD}_j]$, where $\mathit{VD}_i$ and $\mathit{PD}_j$ refer to two different types of 3D lines detected in the RGB-D frame: the former refers to lines that are parallel to at least another 3D line, the latter to lines that are not parallel to any line. The first type of lines can generate vanishing directions $\mathit{VD}_i$ in a single view, which are stored into the graph directly, similarly to planes. In addition, lines that do not have parallel lines detected in this RGB-D frame are marked as potential vanishing direction $\mathit{PD}_j$. In case parallel lines will be detected in successive frames, these lines will also be transferred to $\mathit{VD}_j$, otherwise, they are removed from the E-Graph.

\paragraph{\textbf{Landmarks fusion}.}\label{eg-grow}
For each new input frame we need to extract vectors $\textbf{n}^{\pi}$, $\mathit{VD}$ and $\mathit{PD}$ from the current frame. After rotating $\mathit{VD}_i^c$ to the world coordinate frame as $\mathit{VD}_i^w$, if the direction between $\mathit{VD}_i^c$ is parallel to $\mathit{VD}_k^w, k\in[0,\dots,m]$, where $m$ is the number of vanishing directions saved in E-Graph, $\mathit{VD}_i^c$ is then associated to the graph. To solve the unsure issues~\cite{salaun2016robust} of vanishing directions, we will unify the direction during the association process by using 
\begin{equation}
    \tilde{VD}_i^c =\left\{\begin{array}{cc}
        \mathit{VD}_i^c & (|norm(\mathit{VD}_i^w \cdot \mathit{VD}_k^w)-1|< th_{vd} )  \\
        -\mathit{VD}_i^c & (|norm(\mathit{VD}_i^w \cdot \mathit{VD}_k^w)+1|<th_{vd})
    \end{array}\right.
\end{equation}
where $norm(\cdot)$ shows a dot product between two normalized vectors and $|\cdot|$ is the absolute difference. $th_{vd}$ is a threshold to check the angle distance between two vectors. To include additional graph connections, we also try to associate $\mathit{PD}_j^c$ with $\mathit{VD}_k^w$ and $\mathit{PD}_k^w$. If new pairs can be made at this stage, the associated $\mathit{PD}$ vectors are transferred to the vanishing directions and fused into the graph. 

Since the vanishing direction is independent from translation motion, $\mathit{VD}_{i}^w$, the vanishing direction in the world coordinate can be obtained as
\begin{equation}
    \mathit{VD}_{i}^w = R_{wc}\mathit{VD}_i^c
\end{equation}
where $R_{wc}$ is the rotation motion from the camera coordinate frame to the world coordinate frame.  

In certain indoor scenes, e.g.\ a corridor or hallway, when a robot moves along the wall, an extended planar region is detected across multiple views, with most of these views encompassing no overlap. To address this issue, we extract the normal vector $[n_x^c,n_y^c,n_z^c]$ of the plane in the camera coordinate, which can be fused into the world coordinate in the same way as the vanishing directions.

\paragraph{\textbf{Edge connection}.}
In E-Graph, all landmarks come from keyframes that follow the decision mechanisms of a feature-based SLAM system~\cite{murTRO2015,Li2021PlanarSLAM}, which we summarize in the following. A new keyframe is detected if it satisfies one of the following two conditions: 1) 20 frames have passed from the last keyframe; 2) the current frame tracks less than 85\% points and lines correspondences with the last keyframe.  
Furthermore, when the current frame detects a new plane or a new vanishing direction, the frame is considered as a new keyframe. In addition, new landmarks connected to this keyframe are also merged into the graph at this stage.

By sequentially processing keyframes, if more than two pairs of matched landmarks are observed between two keyframes, an edge will be created to connect the respective two graph nodes. As shown in Figure~\ref{fig:relative_rotation}, $C_j$ and $C_k$ detect the plane $\pmb{\pi}$ and the same vanishing point generated by $L_1$ and $L_2$. Notably, even if these two frames do not have any correspondence, they can still be connected in our E-Graph.

\section{Experiments}
In this section, the proposed system is evaluated on different indoor benchmarks: ICL-NUIM~\cite{handa:etal:ICRA2014} and TUM RGB-D~\cite{sturm2012benchmark}.
ICL-NUIM~\cite{handa:etal:ICRA2014} contains eight synthetic sequences recorded in two scenarios (living room and office room). TUM RGB-D~\cite{sturm2012benchmark} is recorded in real scenarios and includes varied sequences in terms of texture, scene size, presence of moving objects, etc.

\paragraph{\textbf{Rotation estimation}.}
The proposed rotation algorithm is compared with other state-of-the-art orientation estimation approaches. Compass~\cite{kim2018indoor} makes use of a single line and plane. OPRE~\cite{zhou2016divide} and GOME~\cite{joo2016globally} estimate the distribution of surface normal vectors based on depth maps. OLRE~\cite{bazin20123} and ROVE~\cite{lee2015real} take advantage of vanishing directions for rotation estimation. Importantly, Compass, GOME, OLRE, OPRE, and P-SLAM~\cite{Li2021PlanarSLAM} are all based on the MW assumption, while our method, ORB-SLAM2~\cite{mur2015orb} and ROVE are designed for general scenes. 

\paragraph{\textbf{Translation estimation}.}
Since the rotation of the current frame is estimated from a keyframe that may not be overlapping with the current frame, we follow the 3D translation estimation model~\cite{murORB2,Li2021PlanarSLAM} to estimate the translation $\textbf{t}$ based on the predicted rotation. In this module, re-projection errors from point-line-plane feature correspondences are used to build a target optimization function, 
$ \textbf{t} = argmin(\sum_{j=0}^n e^{\pi}_{i,j}\Lambda^{\pi} e^{\pi}_{i,j}+e^{L}_{i,j}\Lambda^{L} e^{L}_{i,j}+ e^{P}_{i,j}\Lambda^{L} e^{P}_{i,j})
$, where $e^{\pi}$, $e^{L}$ and $e^{P}$ are re-projection error functions for planes, lines and points, respectively. The target function is optimized by using the Levenberg-Marquardt method. The translation is compared with the following state-of-the-art methods. ORB-SLAM2~\cite{murORB2} and ORB-SLAM3~\cite{campos2020orb} are popular keypoint-based SLAM systems. In our experiments, for fairness of comparison the loop closure is removed to reduce the effect of the back-ends. SP-SLAM~\cite{zhang2019point} additionally uses points and planes in the tracking and optimization modules based on ORB-SLAM2. P-SLAM~\cite{li2020structure} assumes the indoor environments as MW, and includes a refinement module to make the tracking process more robust. Moreover, we also compare our system with GPU-based methods, including BadSLAM~\cite{8954208} and BundleFusion~\cite{dai2017bundlefusion}. 

\paragraph{\textbf{Dense mapping}.}
In this paper, a mapping module is implemented to reconstruct unknown environments in sparse and dense types. The sparse map is reconstructed by the point-line-plane features extracted from keyframes, which supports a frame-to-map pose refinement step. Since sparse maps cannot provide enough information for robots, our system also generates a dense mesh map incrementally based on CPU. When a new keyframe is generated from the tracking thread, we make use of the estimated camera pose and the RGB-D pair to build a dense TSDF model based on~\cite{zhou2013dense,niessner2013real}. After that, the marching cubes method~\cite{lorensen1987marching} is exploited to extract the surface from voxels.

\paragraph{\textbf{Metrics.}}
The metrics used in our experiments include absolute trajectory error (ATE), absolute rotation error (ARE), and relative pose error (RPE) that shows the difference in relative motion between two pairs of poses to evaluate the tracking process. Our results are reported in Table~\ref{tab:ATE} and obtained on an Intel Core @i7-8700 CPU @3.20GHz and without any use of GPU resources.

\begin{table}
\caption{Comparison of the average value of the absolute rotation error (degrees) on ICL-NUIM and TUM RGB-D structural benchmarks. The best result for each sequence is bolded. $\times$ shows that the method fails to track the orientation. }
\scalebox{0.8}{
\begin{tabular}{l|ccccccccc}
Sequence & Ours & Compass~\cite{kim2018indoor} & OPRE~\cite{zhou2016divide}     & GOME~\cite{joo2016globally}     & ROVE~\cite{lee2015real}     & OLRE~\cite{bazin20123}     & ORB2~\cite{murORB2} & P-SLAM~\cite{Li2021PlanarSLAM} \\ \hline
office room 0     & \textbf{0.11}      & 0.37    & 0.18     & 5.12     & 29.11    & 6.71     & 0.40         &0.57        \\
office room 1     &  \textbf{0.22}    & 0.37    & 0.32     & $\times$ & 34.98    & $\times$ &2.30                 & \textbf{0.22}     \\
office room 2     & 0.39     & 0.38    & 0.33     & 6.67     & 60.54    & 10.91    &0.51              &\textbf{0.29}         \\
office room 3     & 0.24     & 0.38    & \textbf{0.21}     & 5.57     & 10.67    & 3.41     &0.36             &\textbf{0.21}         \\ \hline
living room 0     &0.44      & \textbf{0.31}    & $\times$ & $\times$ & $\times$ & $\times$ &    0.97 &0.36          \\
living room 1     & \textbf{0.24}     & 0.38    & 0.97     & 8.56     & 26.74    & 3.72     &0.22                 & 0.26    \\
living room 2     & 0.36     & \textbf{0.34}    & 0.49     & 8.15     & 39.71    & 4.21     &   0.83        & 0.44        \\
living room 3     & 0.36     & 0.35    & 1.34     & $\times$ & $\times$ & $\times$ &  0.42         &\textbf{0.27}         \\ \hline
$f3\_stru\_notex$    & 4.46    & 1.96    &3.01     & 4.07     & $\times$    & 11.22     &    $\times$       & 4.71          \\
$f3\_stru\_tex$    & \textbf{0.60}    & 2.92    & 3.81     & 4.71     & 13.73   & 8.21     &  0.63         & 2.83 \\
$f3\_l\_cabinet$    & \textbf{1.45}     & 2.04    & 36.34     & 3.74     & 28.41    & 38.12  & 2.79           & 2.55\\ 
$f3\_cabinet$    & 2.47 & 2.48    & 2.42     & 2.59     &$\times$   & $\times$    & 5.45           &\textbf{1.18} \\
\hline
\end{tabular}}\label{tab_are}
\end{table}

\subsection{ICL NUIM dataset}
As shown in Table~\ref{tab_are}, the proposed method outperforms other MW-based and feature-based methods in terms of average rotation error. In \textit{office room} sequences, OPRE and P-SLAM also perform well since orthogonal planar features can be found in the environment. However, in \textit{office room 0}, parts of the camera movement only contain a single plane and some lines, leading to performance degradation, while our method achieves robust orientation tracking by taking advantage of a set of non-parallel planes and lines.    

Furthermore, we compare the translation results against two feature-based methods as shown in Table~\ref{tab:ATE}. The first four sequences are related to a living room scenario, while the remaining sequences are from an office scenario. All methods obtain good results in \textit{living room 0} where the camera moves back and forth between the two parallel walls. P-SLAM detects a good MW model, and ORB-SLAM3 also observes enough features, benefiting from paintings hanging on the wall and small furniture. Compared with the living room, the office room has many low-textured regions. The performance of feature-based algorithms is not as good as in the living room scenes, especially in \textit{office room 1} and \textit{office room 3}.

\begin{table}
\centering
\caption{Comparison in terms of translation RMSE (m) for ICL-NUIM and TUM RGB-D sequences. $\times$ means that the system fails in the tracking process.}
\begin{tabular}{l|ccc}
  Sequence & Ours  &{P-SLAM\cite{Li2021PlanarSLAM}} & {ORB-SLAM3\cite{campos2020orb}}  \\
  \hline
office room 0 &\textbf{0.014}   &0.068   &0.035      \\
office room 1 &\textbf{0.013}  & 0.020  &0.091       \\
office room 2 & 0.020  &0.011    &\textbf{0.010}    \\
office room 3 &\textbf{0.011} &0.012  &0.096   \\ \hline
living room 0 &0.008 &\textbf{0.006}     & \textbf{0.006}     \\
living room 1 &\textbf{0.006}  & 0.015    & 0.206    \\
living room 2 &\textbf{0.017} &0.020     & 0.018     \\
living room 3 &0.021  &\textbf{0.012}    & 0.019   \\ \hline
$f1\_360$ &0.114  &$\times$      & \textbf{0.108}    \\
$f1\_room$ & \textbf{0.095} &$\times$  &$\times$   \\
$f2\_rpy$ & \textbf{0.002} & 0.154 &0.003   \\
$f2\_xyz$ &\textbf{0.003} &0.009 &0.004   \\
$f3\_l\_o\_house$ &0.012 &0.122 &\textbf{0.009}   \\
$f3\_stru\_notex$ &\textbf{0.017}  & 0.025    &$\times$  \\
$f3\_l\_cabinet$ &\textbf{0.058}  & 0.071     &0.072  \\ \hline
\end{tabular}
\label{tab:ATE}
\end{table}

\begin{figure}
    \centering
    \subfigure[ATE]{
    \includegraphics[width=0.95\textwidth]{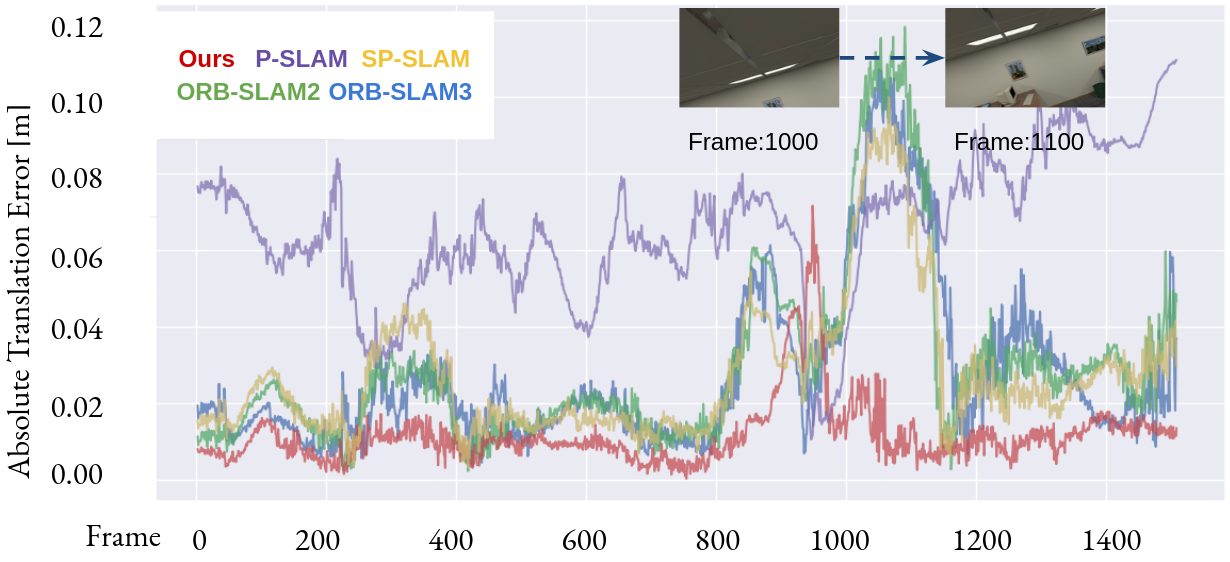}}
    \subfigure[Ours]{
     \includegraphics[width=0.30\textwidth]{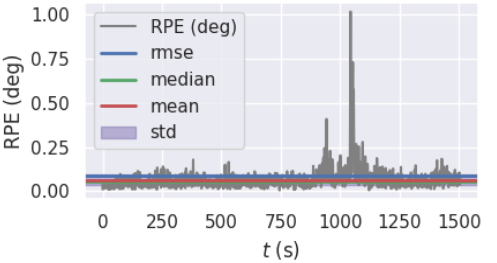}}
     \subfigure[P-SLAM]{
     \includegraphics[width=0.30\textwidth]{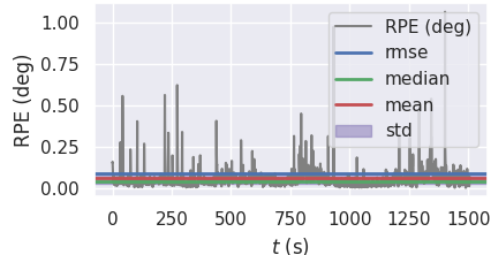}}
     \subfigure[ORB-SLAM3]{
     \includegraphics[width=0.30\textwidth]{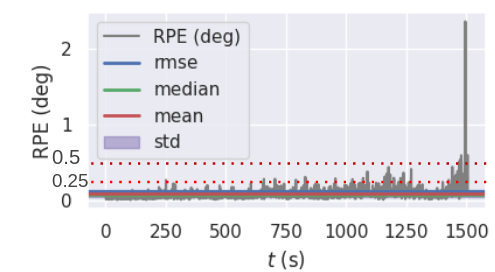}}
    \caption{Comparison of the proposed system against state-of-the-art methods in the \textit{office room 0} sequence of ICL NUIM in terms of mean/average absolute translation errors (top) and rotation errors (bottom). }
    \label{fig:ork0}
\end{figure}

To analyze the relationship between rotation and translation results of different methods, absolute translation and rotation errors on the \textit{office room 0} sequence are presented in Figure~\ref{fig:ork0}. When the camera moves to the ceiling, the number of detected features decreases, then an interesting phenomenon is witnessed (see also Figure~\ref{fig:ork0}(a)): the tracking error of feature-based systems quickly and drastically increases, then gradually fades as the number of features increases. At the same time, our method and P-SLAM exhibit a more robust performance when they face this challenge. An important difference is that, while P-SLAM underperforms due to the non-rigid MW scene, our method's performance is accurate thanks to the use of the E-Graph, which demonstrates to be more flexible than MW-based paradigms.

\begin{figure}
\centering
\begin{minipage}[]{.23\linewidth}
    \centering
    \subfigure[]{ \includegraphics[width=\linewidth]{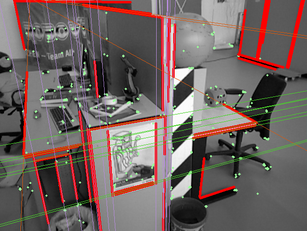}}
    \subfigure[]{\includegraphics[width=\linewidth]{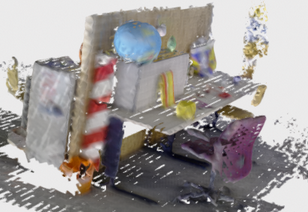}}
\end{minipage} 
\begin{minipage}[]{.5\linewidth}
    \centering
    \subfigure[]{
    \includegraphics[width=1\linewidth]{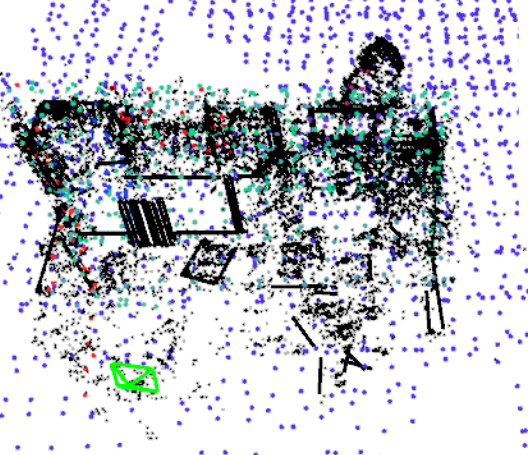}}
\end{minipage}
\begin{minipage}[]{.19\linewidth}
    \centering
    \subfigure[]{ \includegraphics[width=\linewidth]{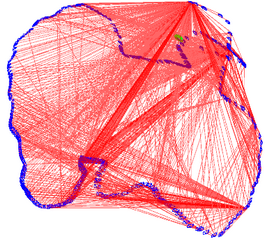}}
    \subfigure[]{\includegraphics[width=\linewidth]{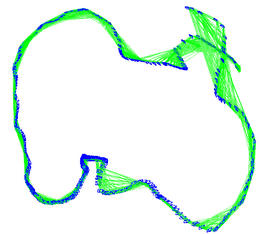}}
\end{minipage} 
\caption{Scene and graphs of \textit{f3\_l\_o\_house}. (a) 2D image, (b) dense mesh model, (c) sparse map, (d) E-Graph, (e) co-visibilitity graph.}
\label{fig:lcabinet}
\end{figure}

\subsection{TUM RGB-D}
Different types of sequences are included from the TUM RGB-D benchmark, which aims to test general indoor scenes with low-textured scenes and sharp rotational motions. \textit{f1\_360}, \textit{f1\_room}, \textit{f2\_rpy} and \textit{f2\_xyz} are recorded in real office scenes, but the camera's rotation motion changes sharply especially in the first sequence. \textit{f3\_l\_o\_house}, \textit{f3\_sn\_near} and \textit{f3\_l\_cabinet} contain more structural information, where \textit{f3\_sn\_near} is built on two white corners, and \textit{f3\_l\_cabinet} records several movements around the white cabinet. Table~\ref{tab_are} shows that ROVE, OLRE and ORB-SLAM2 have problems in low/non-textured regions. In \textit{f3\_l\_cabinet} that is not a rigid MW environment, the quality of depth maps is noisy, the surface normal maps extracted by OPRE have a negative effect on rotation estimation. 

\begin{table}
    \centering
    \caption{ATE RMSE results (cm) on the TUM RGB-D dataset. Results for BundleFusion and BadSLAM are taken from~\cite{8954208}}
    \begin{tabular}{l|cccc}
         \multirow{2}{*}{Sequence}& Ours  & BundleFusion~\cite{dai2017bundlefusion} & ElasticFusion~\cite{whelan2015elasticfusion} &BadSLAM~\cite{8954208}  \\ 
         & CPU  & GPU &GPU &GPU \\ \hline
        f1\_desk & 1.0  & 1.6  &2.0  &1.7  \\
        f2\_xyz &0.7  & 1.1  &1.1  & 1.1\\
        f3\_office &1.4  & 2.2 &3.6  &1.7 \\ \hline 
    \end{tabular}
    \label{tab:my_label}
\end{table}

\begin{figure}
\centering
\begin{minipage}[]{.15\linewidth}
    \centering
    \subfigure[]{ \includegraphics[width=\linewidth]{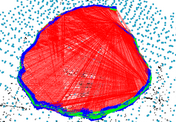}}
    \subfigure[]{\includegraphics[width=\linewidth]{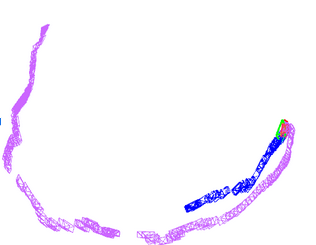}}
\end{minipage} 
\begin{minipage}[]{.28\linewidth}
    \centering
    \subfigure[]{
    \includegraphics[width=1\linewidth]{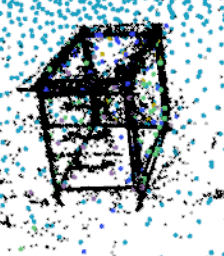}}
\end{minipage}
\begin{minipage}[]{.40\linewidth}
    \centering
    \subfigure[]{ \includegraphics[width=\linewidth]{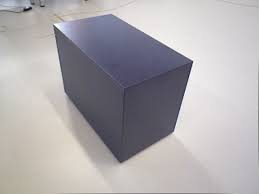}}
\end{minipage} 
\caption{Scene and graph of \textit{f3\_cabinet}. (a) E-Graph, (b) trajectory from ORB-SLAM3, (c) sparse map, (d) 2D image.}
\label{fig:cabinet}
\end{figure}

For structural sequences listed in Table~\ref{tab_are}, P-SLAM shows stable performance. In Table~\ref{tab:ATE}, general scenes are added as a comparison. As listed in Table~\ref{tab:ATE}, the keypoint-based method~\cite{campos2020orb} cannot achieve robust results in \textit{f3\_sn\_near}, i.e.\ , a textureless scenario, while the MW-based method~\cite{Li2021PlanarSLAM} has problems when the scene structure breaks the MW assumption, by reporting a low performance in \textit{f2\_rpy} and \textit{f3\_l\_o\_house}, and even losing track in \textit{f1\_360} and \textit{f1\_room}. Therefore, the proposed method shows more robust performances in different types of scenarios, compared with MW-based systems~\cite{Li2021PlanarSLAM,kim2018indoor} and feature-based approaches~\cite{murORB2,campos2020orb}. Furthermore, compared with GPU-based systems, our system only works on limited computation sources.
As shown in Figure~\ref{fig:lcabinet}, \textit{f3\_l\_o\_house} is used to compare E-Graph and co-visibility graph. As clearly shown, E-Graph allows connecting more distant keyframes than a co-visibility graph. When two keyframes can be connected together, drifting phenomena can more easily be limited, in a similar way to the underlying idea behind loop closure. %
The cabinet scene is also a difficult sequence for point-based methods (see Figure~\ref{fig:cabinet}(b)) since point features are concentrated in a few boundary regions. However, our method can deal with this type of scene where the same plane is observed in a number of frames.


\section{Conclusion}
This paper proposed a new graph structure, E-Graph, to reduce tracking drift based on plane normals and vanishing directions in a scene, which can be used to build a rotation connection between two frames without visual overlap. The advantage of this idea is that rotation errors that occur between two frames have small or no effect on this relative rotation estimation step. Based on the proposed graph, a minimal solution is presented, that shows that two landmarks and two correspondences can be used to solve the relative camera pose. Therefore, the proposed method is better suited for texture-less scenes compared with traditional minimal solutions based on co-visible features. However, the proposed method also has limitations. Compared with point-based systems, our approach requires more types of features. Furthermore, since we need vanishing directions and plane vectors, the method is more suitable for man-made scenes.       

\textbf{Feature work}. The E-Graph is a new tool to establish connections across frames and keyframes. An interesting topic for future exploration is considering a covisibility graph and our graph together to revisit pose estimation and obtain further improvements in drift removal.  

\textbf{Acknowledgements}. We gratefully acknowledge Xin Li, Keisuke Tateno, Nicolas Brasch and Dr. Liang Zhao for the helpful discussion.


\end{document}